*Systems biology*

# Tensor Decomposition with Relational Constraints for Predicting Multiple Types of MicroRNA-disease Associations

Feng Huang[1,†], Xiang Yue[2,†], Zhankun Xiong[1], Zhouxin Yu[1] and Wen Zhang[1,*]

[1]College of Informatics, Huazhong Agriculture University, Wuhan, Hubei, China, [2]Department of Computer Science & Engineering, The Ohio State University, Columbus, OH, USA

*To whom correspondence should be addressed.

†Equal contributions.

**Abstract**

**Motivation:** MicroRNAs (miRNAs) play crucial roles in multifarious biological processes associated with human diseases. Identifying potential miRNA-disease associations contributes to understanding the molecular mechanisms of miRNA-related diseases. Most of the existing computational methods mainly focus on predicting whether a miRNA-disease association exists or not. However, the roles of miRNAs in diseases are prominently diverged, for instance, *Genetic* variants of microRNA (mir-15) may affect expression level of miRNAs leading to B cell chronic lymphocytic leukemia, while *circulating* miRNAs (including mir-1246, mir-1307-3p, etc.) have potentials to detecting breast cancer in the early stage.
**Results:** In this paper, we aim to predict multi-type miRNA-disease associations instead of taking them as binary. To this end, we innovatively represent *miRNA-disease-type* triplets as a tensor and introduce Tensor Decomposition methods to solve the prediction task. Experimental results on two widely-adopted miRNA-disease datasets: HMDD v2.0 and HMDD v3.2 show that tensor decomposition methods improve a recent baseline in a large scale (up to 38% in top-1 F1). We further propose a novel method, Tensor Decomposition with Relational Constraints (TDRC), which incorporates biological features as relational constraints to further the existing tensor decomposition methods. Compared with two existing tensor decomposition methods, TDRC can produce better performance while being more efficient.
**Availability**: https://github.com/BioMedicalBigDataMiningLab/TDRC
**Contact**: zhangwen@mail.hzau.edu.cn
**Supplementary information**: Supplementary Materials can be founded in the submission files.

## 1 Introduction

MicroRNAs (miRNAs) are a kind of small non-coding RNAs containing about 22 nucleotides (Bartel, 2004), which play crucial roles in multifarious biological processes, such as cell growth (Karp and Ambros, 2005), tissue differentiation (Miska, 2005), cell proliferation (Cheng, et al., 2005), embryonic development and apoptosis (Lu, et al., 2008; Xu, et al., 2004). Aberrant expressions of miRNAs are associated with a broad range of human diseases (Jiang, et al., 2009). For example, miR-129, miR-142-5p, and miR-25 are differentially expressed between pediatric central nervous system neoplasms and normal tissue, indicating their role in oncogenesis (Sredni, et al., 2011). Therefore, identifying miRNA-disease associations contributes to understanding the molecular mechanisms of miRNA-related diseases, improving the discovery of potential biomarkers (Jones, et al., 2014) and helping develop novel therapies (Li, et al., 2009).

In earlier studies, wet experiments including reverse transcription PCR (Siebert, 1999) and northern blotting (Varallyay, et al., 2008) were adopted to identify miRNA-disease associations. However, such wet experiments were often time-consuming, laborious and costly.

Recently, a surge of computational methods has been proposed to overcome the limitations of traditional laboratory experiments based on publicly available miRNA-disease databases (Huang, et al., 2019; Jiang, et al., 2009; Li, et al., 2014; Yang, et al., 2010). For example, (Xuan, et al., 2015) and (Liu, et al., 2017) exploit random walk-based methods to predict potential miRNA-disease associations. (Xiao, et al., 2017) and Chen et al. (Chen, et al., 2020; Chen, et al., 2018) applied matrix factorization models to complete the miRNA-disease association matrix so as to identify potential associations. Zeng et al. (Zeng, et al., 2018) developed a structural perturbation method on the miRNA-disease bilayer network to infer potential links between miRNAs and diseases. Peng et al. (Peng, et al., 2019) proposed a deep learning framework for miRNA-disease associations identification. Zhang et al. (Zhang, et al., 2019) built a similarity-based framework based on known miRNA-disease associations. Huang et al. (Huang, et al., 2019) performed systematic comparison among 36 readily available prediction methods and provided benchmarking results of all predictors.

However, all the aforementioned methods mainly focus on predicting whether a miRNA-disease association exists or not. As can be seen from Figure 1, there are multiple types of miRNA-disease associations, and the roles of miRNAs in diseases are prominently diverged and the molecular mechanism of miRNA-related human diseases are various (Huang, et al., 2019). For example, genetic variants in miRNA genes may affect the expression level of miRNAs leading to diseases: B cell chronic lymphocytic leukemia may result from down-regulation of the mir-15 and the mir-16 induced by the deletion of chromosome 13q14 (Calin, et al., 2002). Besides, circulating miRNAs have potentials to assist clinical diagnosis. For example, (Shimomura, et al., 2016) reported a novel combination of serum miRNAs (including mir-1246, mir-1307-3p, mir-4634, mir-6861-5p and mir-6875-5p) for detecting breast cancer in the early stage due to the significant expression difference between the patients and the normal. Thus, it is difficult, even not possible, to fully understand the pathogenesis of diseases implicated with the dysregulations of miRNAs only through exploring the existence of miRNA-disease associations but without knowing their specific types.

Chen et al. (Chen, et al., 2015) made the first attempt to predict multiple types of miRNA-disease associations using the restricted Boltzmann machine, however, the model merely made use of known multiple types of miRNA-disease associations and did not consider the relationships of miRNA-miRNA pairs or disease-disease pairs. Recently, Zhang et al. (Zhang, et al., 2018) proposed the NLPMMDA method, which employed label propagation on miRNA similarity network and disease similarity network to propagate label information of each type of miRNA-disease associations in turns. However, NLPMMDA treats predicting one type of miRNA-disease as an independent task and therefore ignores the correlations between association types.

Tensor is a multidimensional array, as an extended concept of matrix. 3-dimensional tensors are commonly used for triplet data analysis such as multi-relational networks (Nickel, et al., 2011), recommendation system (Rendle, et al., 2009), and knowledge graph (Trouillon, et al., 2017). A miRNA-disease-type can be naturally modeled as a binary tensor where every element represents whether the corresponding entry (miRNA, disease, type) exists or not. Our goal is to complete the tensor for exploring the unobserved triple associations. Tensor decomposition (Kolda and Bader, 2009) is a popular method for tensor completion by decomposing a tensor as the product of several small tensors to obtain its approximation. Besides, it also can capture the complicated multilinear relationship between miRNAs, diseases and association types through the tensor multiplications to overcome the aforementioned limitations.

In this paper, we represent the multi-type miRNA-disease associations as a tensor and formulate the multi-type miRNA-disease association prediction as a tensor completion task. We implement several classic tensor decomposition models and investigate their effectiveness on predicting multiple types of miRNA-disease associations. Our experimental results on two popular datasets HMDD v2.0 and HMDD v3.0 demonstrate that formulating the data as tensors is effective to capture complex relations between miRNAs and diseases. By comparing with a recent baseline NLPMMDA (Zhang, et al., 2018), the tensor decomposition methods improve up to 38% on top1@F1 score. Further, we observe that some of the existing tensor decomposition methods only utilize the association information, which may fail to deeply capture correlations between miRNAs and diseases. Inspired by the previous studies that incorporate auxiliary information to improve the biological link predictions (Zhang, et al., 2018), we propose a novel tensor decomposition-based model, named as tensor decomposition with relational constraints (TDRC), which incorporates the information of miRNA-miRNA similarity and disease-disease similarity as decomposition constraints. Experimental results show that TDRC further improves the performance by comparing with existing tensor decomposition methods.

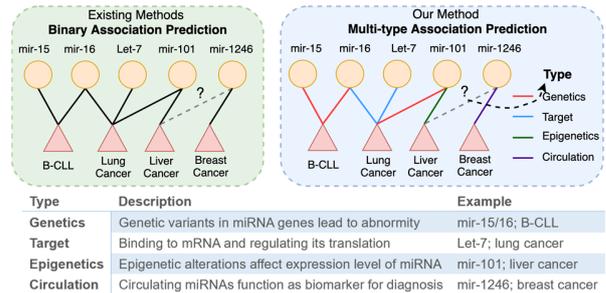

**Fig. 1. The motivation of our study and previous studies as well as descriptions and examples of four miRNA-disease association types from HMDD 2.0 database.**

Our contributions are summarized as follows:

- We study the multi-type miRNA-disease associations prediction task, on which few attentions have been paid previously. We introduce tensor decomposition methods to solve it and achieve decent improvement compared with a recent baseline.
- We investigate the effectiveness of existing tensor decomposition-based models on this task and propose TDRC to further the performance, which integrates auxiliary biological information as constraints
- Theoretically, we provide a high-efficiency optimization algorithm for our proposed model TDRC in virtue of the alternating direction method of multipliers (ADMM) framework, and resort the conjugate gradient (CG) method to avoid computing an inverse matrix in inner iterations of ADMM for lower time complexity.
- The experimental results reveal that TDRC can produce robust and satisfying performance meanwhile being efficient.

## 2 Materials

### 2.1 Datasets

The Human MiRNA Disease Database (HMDD) is a database that contains experimentally verified human miRNA- disease associations. HMDD v2.0 (Li, et al., 2014) serves as a comprehensive data source in many miRNA-related computational researches (Chen, et al., 2015; Zhang, et al., 2018). The curated associations are classified into four types based on the evidence from *genetics, epigenetics, circulation* and *miRNA-target interactions*. Recently, HMDD v3.0 (Huang, et al., 2019) has been released (we used the latest version v3.2, released on March 27, 2019).

**Table 1.** Statistics of the data used in the study (Density = $\frac{\#Associations}{\#miRNA*\#Disease}$)

| dataset | #miRNA | #Disease | #Type | #Associations | Density |
| --- | --- | --- | --- | --- | --- |
| v2.0 | 324 | 169 | 4 | 1492 | 0.681% |
| v3.2 | 713 | 447 | 5 | 16341 | 1.025% |

Comparing to HMDD v2.0, this new version covers more entries, and provides 2 more types (tissue and other). We downloaded five types of miRNA-disease associations (the associations in the "other" category are not able to be downloaded with some unknown problem). To get dense data, we remove those miRNAs (diseases) that involve less than two associations in total across all types. We obtain 16341 multi-type miRNA-disease associations, involving 713 miRNAs and 447 diseases in HMDD v3.2 and 1492 associations, involving 324 miRNAs and 169 diseases in HMDD v2.0. The statistics of two datasets are shown in Table 1. We also show the number of miRNA-disease associations in each type of two datasets in the Supplementary Figure S1.

Besides, we downloaded disease descriptors from Medical Subject Heading (MeSH). MeSH is a comprehensive controlled vocabulary thesaurus about life science for the facility of searching. Disease descriptors can be used to calculate disease semantic similarity (Wang, et al., 2010).

### 2.2 Problem description

Given a set of miRNAs $\mathcal{E} = \{e_1, e_2, ..., e_m\}$, a set of diseases $\mathcal{D} = \{d_1, d_2, ..., d_n\}$ and a set of association types $\mathcal{R} = \{r_1, r_2, ..., r_t\}$, we can construct a multi-relation bipartite graph $\mathcal{G}$. A triple $(e_i, r_k, d_j)$ as a link in the graph $\mathcal{G}$ denotes that the association between the miRNA $e_i$ and the disease $d_j$ with the type $r_k$. Despite some associations have been retrieved and classified, there are still a large number of associations unverified. Inferring those potential links in the graph $\mathcal{G}$ can enhance our comprehension of the underlying pathogenesis of diseases at the molecular level of miRNAs. To this end, we aim to identify the potential associations in the graph $\mathcal{G}$ with explicitly knowing the specific type of each potential association.

The above-constructed graph $\mathcal{G}$ can naturally be organized as a binary three-way tensor $\mathcal{X} \in \{0,1\}^{m \times n \times t}$ with miRNA mode, disease mode and type mode, where each slice is the adjacency matrix with regard to a type of miRNA-disease associations. More specifically, an entry $x_{ijk}$ of the tensor is set to 1 if $(e_i, r_k, d_j) \in \mathcal{G}$. Otherwise, the entries are set to 0. The tensor $\mathcal{X}$ is extremely sparse with many unknown entries and thus it is challenging to reach the goal only by using known links. Hence, we consider biological similarities as auxiliary information to tackle the challenge.

### 2.3 Auxiliary information

According to MeSH descriptors, the hierarchical relationships of diseases can be transformed into Directed Acyclic Graphs (DAGs), where nodes represent the diseases and edges represent the relationships between different diseases. As described in (Wang, et al., 2010), DAGs can be used to calculate disease semantic similarity. For a disease $d$, a DAG denoted as $DAG(d) = (N(d), E(d))$ is constructed, where $N(d)$ is the set of all ancestors of $d$ (including itself) and $E(d)$ is the set of links from ancestor disease to their children. The semantic contribution of disease $d_i \in N(d)$ to disease $d$ can be calculated as:

$$C(d, d_i) = \begin{cases} 1 & if\ d_i = d \\ max\{\Delta * C(d, d_j) | d_j \in children\ of\ d_i\} & if\ d_i \neq d \end{cases}$$

where $\Delta$ is the semantic contribution factor, and we set $\Delta = 0.5$ in this work. Then the semantic value of disease $d$ is defined as:

$$SV(d) = \sum_{d_i \in N(d)} C(d, d_i)$$

Finally, the semantic similarity between two diseases $d_i$ and $d_j$ is calculated by:

$$s_{ij}^n = S_{disease}(d_i, d_j) = \frac{\sum_{d \in N(d_i) \cap N(d_j)} (C(d_i, d) + C(d_j, d))}{SV(d_i) + SV(d_j)}$$

Moreover, we resort to the measure mentioned in (Xiao, et al., 2017) to calculate the miRNA functional similarity between two miRNAs $e_i$ and $e_j$ as follows:

$$s_{ij}^m = S_{miRNA}(e_i, e_j) = \frac{\sum_{d \in \mathcal{D}(e_i)} S_{disease}(d, d_j^*) + \sum_{d \in \mathcal{D}(e_j)} S_{disease}(d, d_i^*)}{|\mathcal{D}(e_i)| + |\mathcal{D}(e_j)|}$$

where $\mathcal{D}(e_i)$ represents the set of diseases that are associated with $e_i$ in at least one association type, $|\mathcal{D}(e_i)|$ is the number of elements in the set $\mathcal{D}(e_i)$ and $d_i^* = \underset{d_i \in \mathcal{D}(e_i)}{argmax}\ S_{disease}(d, d_i)$.

We denote $\mathbf{S_m} \in \mathbb{R}^{m \times m}$ as the miRNA-miRNA functional similarity matrix with $s_{ij}^m$ as its $(i,j)$th element and $\mathbf{S_n} \in \mathbb{R}^{n \times n}$ as the disease-disease semantic similarity matrix with $s_{ij}^n$ as its $(i,j)$th element.

## 3 Methods

### 3.1 Classic tensor decomposition-based methods

In this section, we first introduce the most commonly-used CANDECOMP/PARAFAC (CP) decomposition. Then, we describe our TDRC method that incorporates biological auxiliary information into the CP Decomposition framework. We finally develop an efficient optimization method for solving TDRC objective function.

#### 3.1.1 CP decomposition

CANDECOMP/PARAFAC (CP) decomposition (Kolda and Bader, 2009) is one of the most common tensor decomposition forms. Given the miRNA-disease-type tensor $\mathcal{X}$, the CP decomposition model can be represented as the following optimization problem:

$$\underset{C,P,F}{min} \|\mathcal{X} - [\![C, P, F]\!]\|^2 \quad (1)$$

Here, $\|\cdot\|$ is the norm of a tensor. $C \in \mathbb{R}^{m \times r}$, $P \in \mathbb{R}^{n \times r}$ and $F \in \mathbb{R}^{t \times r}$ are the factor matrices with respect to the miRNA, disease and type mode, which are usually considered as latent representations for the corresponding modes. $[\![C, P, F]\!]$ is the reconstructed tensor and its $(i,j,k)$th element is calculated by $\sum_l^r c_{il} p_{jl} f_{kl}$, where $c_{il}$, $p_{jl}$ and $f_{kl}$ denote respectively the $(i,l)$th element of $C$, the $(j,l)$th element of $P$ and the $(k,l)$th element of $F$. We call $r$ as the rank of the approximated tensor $[\![C, P, F]\!]$. In general, $r$ is set much lower than $min(m,n)$ so that the low-rank property of the latent representations is enforced. More detailed information for tensor algebra is referred to Supplementary Section 1. The optimization problem (1) can be readily solved by alternating least squares (ALS) method (Kolda and Bader, 2009).

### 3.2 Tensor decomposition with relational constraints
#### 3.2.1 Formulation

The standard CP model only utilizes the association information. We further propose tensor decomposition with relational constraints (TDRC) method to incorporate biological similarities as constraints into the CP model. The whole model architecture of TDRC is illustrated in Fig. 2.

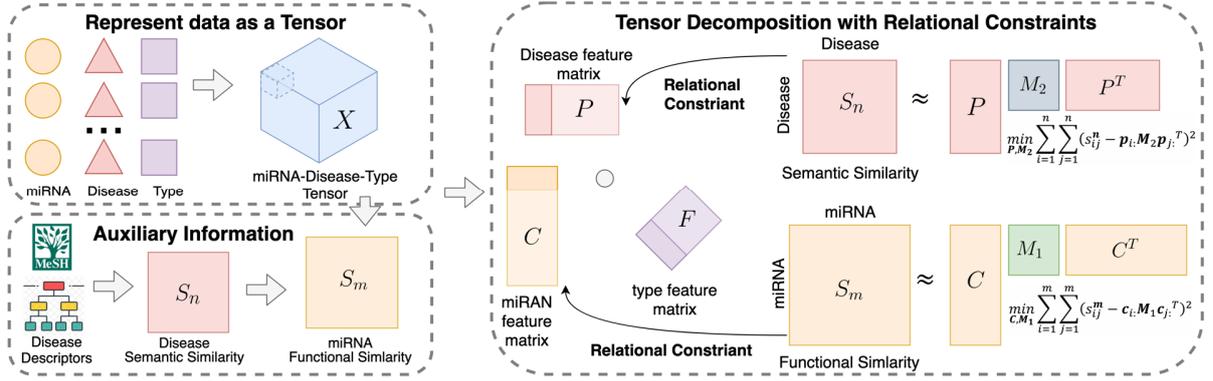

Fig. 2. The illustration of representing data as a tensor and the proposed Tensor Decomposition with Relational Constraints (TDRC) method.

Note the factor matrices $C \in \mathbb{R}^{m \times r}$ and $P \in \mathbb{R}^{n \times r}$ in the CP model share a common latent space with dimensionality of rank $r$, where the $i$th miRNA is encoded as a vector $c_{i:}$ and the $j$th disease is represented as a vector $p_{j:}$. Inspired by (Nickel, et al., 2011), we use a real-valued function $f(x, y) = xMy^T$ to approximate the similarity between two miRNAs (or diseases) for high-quality relational learning, where $M$ is a projection matrix, $x$ and $y$ are the row vectors of $C$ (or $P$). Since we have two different kinds of auxiliary information, the approximation errors are minimized by:

$$\min_{C,P,M} \alpha \sum_{i=1}^{m}\sum_{j=1}^{m}(s_{ij}^m - c_{i:}M_1 c_{j:}^T)^2 + \beta \sum_{i=1}^{n}\sum_{j=1}^{n}(s_{ij}^n - p_{i:}M_2 p_{j:}^T)^2 \quad (2)$$

where $\alpha$ and $\beta$ control the impact of auxiliary information in each part. Different projection matrices $M_1$ and $M_2$ ensure distinct functions to capture various auxiliary information. $s_{ij}^m$ and $s_{ij}^n$ are precomputed in Section 2.3.

Furtherly, the optimization problem (2) can be reformulated in matrix form:

$$\min_{C,P,M} \alpha \|S_m - CM_1 C^T\|_F^2 + \beta \|S_n - PM_2 P^T\|_F^2 \quad (3)$$

where $\|\cdot\|_F$ is the Frobenius norm of a matrix.

The detailed analysis of the relational constraints Eq. (3) is provided in Supplementary Section 2. To ensure the smoothness of the relational learning functions, we further introduce $l_2$ regularization on the projection matrices. By combining Eq. (1), Eq. (3) and $l_2$ regularization terms, we can obtain the following objective function of TDRC:

$$\min_{C,P,F,M} \frac{1}{2}\|\mathcal{X} - [\![C, P, F]\!]\|^2 + \frac{\lambda}{2}(\|M_1\|_F^2 + \|M_2\|_F^2) \\ + \frac{\alpha}{2}\|S_m - CM_1 C^T\|_F^2 \\ + \frac{\beta}{2}\|S_n - PM_2 P^T\|_F^2 \quad (4)$$

where $\lambda$ is the $l_2$ regularization coefficient.

### 3.2.2 Optimization

In this section, we develop an alternately updating rule for optimizing the objective function (4).

*Updating the factor matrix $F$*

Clearly, when the other variables are fixed, the subproblem w.r.t. $F$ is identical to solving the following objective function (Kolda and Bader, 2009):

$$\min_F \frac{1}{2}\|\mathcal{X}_{(3)} - F(C \odot P)^T\|_F^2 \quad (5)$$

where $\mathcal{X}_{(3)}$ is the mode-3 matricization of tensor $\mathcal{X}$ and $\odot$ denotes the Khatri-Rao product (see Definition 2 in Supplementary Section 1). The closed-form solution for $F$ is given by:

$$F = \mathcal{X}_{(3)}(C \odot P)\big((C \odot P)^T(C \odot P)\big)^{-1} \quad (6)$$

*Updating the factor matrices $C$*

The factor matrix $C$ can be obtained through solving the objective function:

$$\min_C \frac{1}{2}\|\mathcal{X}_{(1)} - C(P \odot F)^T\|_F^2 + \frac{\alpha}{2}\|S_m - CM_1 C^T\|_F^2 \quad (7)$$

where $\mathcal{X}_{(1)}$ is the mode-1 matricization of tensor $\mathcal{X}$. Given that Eq. (7) is a fourth-order function of $C$, it is difficult to solve it directly. Inspired by (Kang, et al., 2019), we convert it to an equivalent equality-constrained problem by introducing an auxiliary variable:

$$\min_{C,J_1} \frac{1}{2}\|\mathcal{X}_{(1)} - C(P \odot F)^T\|_F^2 + \frac{\alpha}{2}\|S_m - CM_1 J_1^T\|_F^2 \\ s.t. \quad J_1 = C \quad (8)$$

where $J_1$ is an auxiliary variable. And then we resort to the alternating direction method of multipliers (ADMM) (Boyd, et al., 2011) to solve the problem (8).

First, the augmented Lagrangian function of (8) is introduced as:

$$L(C, J_1, Y_1) = \frac{1}{2}\|\mathcal{X}_{(1)} - C(P \odot F)^T\|_F^2 + \frac{\alpha}{2}\|S_m - CM_1 J_1^T\|_F^2 \\ + tr(Y_1^T(C - J_1)) + \frac{\rho_1}{2}\|C - J_1\|_F^2$$

where $\rho_1 > 0$ is called as the penalty parameter and $Y_1$ is the Lagrange multiplier.

Then, $J_1$ and $C$ can be solved by setting the corresponding first partial derivative to zero. We have:

$$J_1 = (\alpha S_m CM_1 + \rho_1 C + Y_1)(\alpha (CM_1)^T(CM_1) + \rho_1 I)^{-1} \\ C = (\mathcal{X}_{(1)}G + \alpha S_m Q^T + \rho_1 J_1 - Y_1)(G^T G + \alpha QQ^T + \rho_1 I)^{-1} \quad (9)$$

where $G = P \odot F$ and $Q = M_1 J_1^T$; $I$ is an identity matrix of size $(r \times r)$.

Finally, the Lagrange multiplier and the penalty parameter are updated as follows:

$$Y_1 = Y_1 + \rho_1(C - J_1) \\ \rho_1 = \mu \rho_1 \quad (10)$$

where $\mu > 1$ is a given parameter.

*Updating the factor matrix* $P$

The subproblem *w.r.t.* the factor matrix $P$ shares a similar optimization structure as (7). Therefore, $P$ can be updated in the same way:

$$\begin{aligned} J_2 &= (\beta S_d P M_1 + \rho_2 P + Y_2)(\beta (P M_1)^T (P M_1) + \rho_2 I)^{-1} \\ P &= (\mathcal{X}_{(2)} R + \beta S_d Z^T + \rho_2 J_2 - Y_2)(R^T R + \beta Z Z^T + \rho_2 I)^{-1} \\ Y_2 &= Y_2 + \rho_2 (P - J_2) \\ \rho_2 &= \mu \rho_2 \end{aligned} \quad (11)$$

where $R = C \circledast F$, $Z = M_2 J_2^T$ and $\mathcal{X}_{(2)}$ is the mode-2 matricization of tensor $\mathcal{X}$.

*Updating the projection matrices* $M_1$ *and* $M_2$

Considering that the objective functions *w.r.t.* $M_1$ and $M_2$ keep the analogous structure, we unify them into the following generalized optimization problem *w.r.t.* $X$:

$$\min_X \frac{\nu}{2}\|O - UXV^T\|_F^2 + \frac{\lambda}{2}\|X\|_F^2 \quad (12)$$

Inspired by (Yu, et al., 2014), we can exploit the structure in (12) to develop an efficient solver via the conjugate gradient (CG) method. We follow the iterative procedure of CG (Hestenes and Stiefel, 1952), and remold the updating rule in each iteration. Simply, the $k$th iteration of the CG procedure is given by:

$$\begin{aligned} \alpha^{(k)} &= \frac{\|R^{(k)}\|_F^2}{\nu \|UB^{(k)}V^T\|_F^2 + \lambda \|B^{(k)}\|_F^2} \\ X^{(k+1)} &= X^{(k)} + \alpha^{(k)} B^{(k)} \\ R^{(k+1)} &= R^{(k)} - \alpha^{(k)}(\nu U^T U B^{(k)} V^T V + \lambda B^{(k)}) \\ \beta^{(k)} &= \frac{\|R^{(k+1)}\|_F^2}{\|R^{(k)}\|_F^2} \\ B^{(k+1)} &= R^{(k+1)} + \beta^{(k)} B^{(k)} \end{aligned} \quad (13)$$

with $X^{(0)} = 0$, $R^{(0)} = \nu U^T O V - \nu U^T U X^{(0)} V^T V - \lambda X^{(0)}$ and $B^{(0)} = R^{(0)}$. Detailed description, deduction and analysis of the remolded CG procedure are referred to the Supplementary Section 3. By repeating the iterative process until the $\|R^{(k)}\|_F^2$ is small enough, we can obtain an approximate solution of (12) represented as CG($O$, $U$, $V$, $\nu$, $\lambda$), and then we can gain $M_1$ and $M_2$ by CG($S_m$, $C$, $C$, $\alpha$, $\lambda$) and CG($S_n$, $P$, $P$, $\beta$, $\lambda$).

*Optimization algorithm*

According to the above alternately updating rules, the algorithm for solving the optimization problem (4) is summarized in Algorithm 1

## 4 Experiments
### 4.1 Experimental settings

To comprehensively investigate the performances of models in predicting multiple types of miRNA-disease associations, we consider two types of 10-fold cross validation.

- CV$_{type}$: we randomly divide all miRNA-disease pairs that involve at least one type of association into 10 equal-sized subsets. In each turn, one subset is used for testing and the remaining 9 subsets are served as the training set. For each miRNA-disease pair in the testing set, we rank the predictions for all association types, and calculate the top-1 precision, recall and F1 measure. This setting tests how accurate the models can predict the specific type information.

- CV$_{triplet}$: we randomly divide all known miRNA-disease-type triplets into 10 equal-sized subsets. One subset is served as the testing set in turn and the rest associations are used for training. In each fold, we randomly select a fraction of unknown triplets as the negative sample set with the equal size as the testing set. We calculate the area under the precision-recall curve (AUPR), the area under the ROC curve (AUC), and F1-measure. This setting tests how accurate the models can predict the association, which is similar to the experiment settings in previous studies (Chen, et al., 2015) (Zhang, et al., 2018).

Since we are more interested in the type prediction, we treat the CV$_{type}$ as our primary experimental setting. Our implementations of TDRC are based on the tensor learning tool "tensorly" (Kossaifi, et al., 2019).

---

**Algorithm 1** Algorithm for solving problem (4)

---

**Input:** known miRNA-disease-type ternary association tensor $\mathcal{X}$; miRNA similarity matrix $S_m$; disease similarity matrix $S_n$; $\alpha$, $\beta$, $\lambda$ and $r$

**Output:** the factor matrices $C$, $P$ and $F$

**Initialization** $C$, $P$ and $F$ are initialized randomly; $Y_1 = Y_2 = 0$, $\rho_1 = \rho_2 = 1$, $\mu = 1.1$

**Repeat**

    1: Update $M_1$ and $M_2$ using
        $M_1 = $ CG($S_m$, $C$, $C$, $\alpha$, $\lambda$);
        $M_2 = $ CG($S_n$, $P$, $P$, $\beta$, $\lambda$);
    2: Update $F$ via (6)
    3: Update $J_1$, $C$, $Y_1$ and $\rho_1$ as in (9), (10)
    4: Update $J_2$, $P$, $Y_2$ and $\rho_2$ as in (11)

**Until convergence**

---

**CG($O$, $U$, $V$, $\nu$, $\lambda$)** for solving the problem (12)

---

**Initialization** $X^{(0)} = 0$, $R^{(0)} = \nu U^T O V - \nu U^T U X^{(0)} V^T V - \lambda X^{(0)}$

**For** $k = 0, 1, 2, \cdots$

    Update $X$, $R$ and $B$ as in (13)

**Until** $\|R^{(k)}\|_F^2$ is small enough

**output** $X$

---

### 4.2 Parameters analysis

We first analyze how hyperparameters influence the performance of TDRC. TDRC has four hyperparameters: $r$ is the rank of the reconstructed tensor, $\alpha$ and $\beta$ respectively control the contributions of miRNA-miRNA functional similarity and disease-disease semantic similarity, and $\lambda$ is the regularization coefficient. Compared with the other three parameters, $\lambda$ has less influence on the performance. We empirically set $\lambda = 0.001$ and analyze the influence of the other parameters on performance of TDRC. Traversing $r$ within $[2, 4, 6, 8, 10]$ and $\alpha, \beta$ both in the range $[2^{-3}, 2^{-2}, 2^{-1}, 2^0, 2^1]$, we build TDRC models by using grid-search of these parameters and evaluate all methods under 10-CV$_{type}$ setting on both HMDD v3.2 and HMDD v2.0 datasets.

As a result, TDRC produces the best performance (in terms of top-1 precision) on HMDD v3.2 when $r = 4$, $\alpha = 2.0$ and $\beta = 0.125$, and on HMDD v2.0 when $r = 4$, $\alpha = 0.125$ and $\beta = 0.25$.

So we first fix $r = 4$, and investigate the influence of $\alpha$ and $\beta$. As shown in Fig. 3 (a) and (c), in general, a large value of $\alpha$ and a small value of $\beta$ lead to a better performance, which may indicate that the miRNA functional similarity contributes more to capturing the complex relation type information compared with disease semantic similarity.

Then, we fix $\alpha = 2.0$ and $\beta = 0.125$ for HMDD v3.2 dataset, and fix $\alpha = 0.125$ and $\beta = 0.25$ for HMDD v2.0 dataset, and test the impact of $r$. As illustrated in Fig. 3 (b) and (d), the performance of TDRC reaches a peak with a slightly small $r$. It suggests that the low-rank property of the reconstructed tensor may help to further improve the performance.

## 4.2 Comparison experiments

### 4.3.1 Performance

As previously mentioned, most existing methods only focus on predicting binary associations of miRNA-disease. To our best knowledge, only two methods (Chen, et al., 2015) (Zhang, et al., 2018) are designed for the goal of predicting the multiple types of miRNA-disease associations, and we select the latest one (NLPMMDA, by (Zhang, et al., 2018)) as a baseline method. NLPMMDA independently predicted each type of miRNA-disease associations by label propagation on miRNA-miRNA similarity network and disease-disease similarity network. NLPMMDA has two free parameters: $\lambda_d$ and $\lambda_m$. We set $\lambda_d = \lambda_m = 0.2$ in NLPMMDA for both HMDD v2.0 and HMDD v3.2 as the same as the original paper.

We also include two tensor decomposition methods, CP (Kolda and Bader, 2009) and TFAI (Narita, et al., 2012) as baselines. CP is the standard tensor decomposition method as we introduced in Section 3.1. TFAI considers incorporating auxiliary information into CP model via introducing graph Laplacian regularizations. For fairness, we uniformly set the rank of the reconstructed tensor in CP and TFAI as $r = 4$. Since the Laplacian regularization coefficients also control the contributions of miRNA-miRNA functional similarity and disease-disease semantic similarity in TFAI, we set them as $\alpha = 2.0$, $\beta = 0.125$ for HMDD v3.2 and $\alpha = 0.125$, $\beta = 0.25$ for HMDD v2.0 as the same as the settings in TDRC. Besides, they use the same tensor learning tool and convergence criteria as in TDRC.

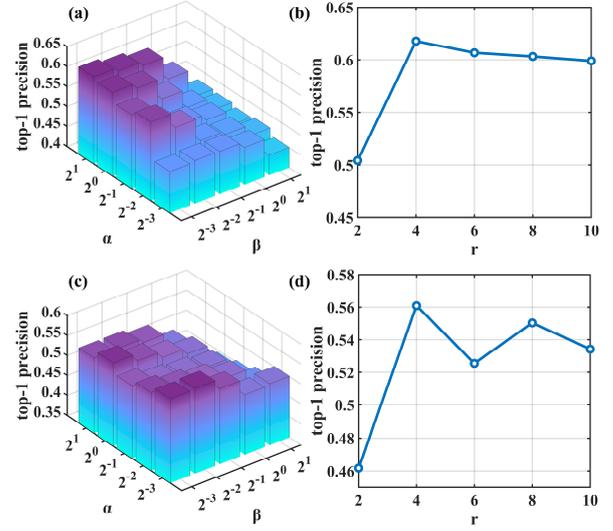

**Fig. 3.** Impact of hyperparameters $\alpha$, $\beta$ of TDRC based on HMDD v3.2 (a), HMDD v2.0 (c) and the influence of $r$ based on HMDD v3.2 (b), HMDD v2.0 (d).

As shown in Table 2, TDRC obviously performs better than others both on HMDD v3.2 and HMDD v2.0 under $CV_{type}$ setting. Compared with the heterogeneous network-based method NLPMMDA, tensor decomposition-based methods make significant improvement. To be more specific, the three measures (top-1 precision, top-1 recall and top-1 f1) of CP, TFAI and TDRC are almost 2 times higher than NLPMMDA on average on HMDD v3.2. The reason for the gap between NLPMMDA and other CP-based models is that NLPMMDA only models the binary associations of miRNA-disease pairs related with every association type but CP-based models dissect the data in a higher dimensional perspective through the tensor decomposition and capture complicated ternary relationships of miRNA-disease-type triples. One can also observe that TFAI works slightly worse on HDMM v3.2 than CP, which may be attributed to its weak ability of incorporating with the auxiliary information. In contrast, the proposed method TDRC achieves much better performance than TFAI, and it is reasonably believed that TDRC makes more sufficient use of the auxiliary information. Further, some similar conclusions can be drawn under 10-$CV_{triplet}$ as shown in Table 3.

**Table 2.** The performances of all methods Evaluated by 10-$CV_{type}$ on HMDD v3.2 and HMDD v2.0

| method | HMDD v3.2 | | | HMDD v2.0 | | |
| --- | --- | --- | --- | --- | --- | --- |
| | Top-1 precision | Top-1 recall | Top-1 f1 | Top-1 precision | Top-1 recall | Top-1 f1 |
| NLPMMDA | 0.1844 | 0.1380 | 0.1579 | 0.4397 | 0.3919 | 0.4144 |
| CP | 0.5956 | 0.4539 | 0.5152 | 0.5543 | 0.4942 | 0.5225 |
| TFAI | 0.5888 | 0.4518 | 0.5113 | 0.5410 | 0.4826 | 0.5101 |
| **TDRC** | **0.6178** | **0.4741** | **0.5365** | **0.5609** | **0.4999** | **0.5286** |

**Table 3.** The performances of all methods evaluated by 10-$CV_{triplet}$ on HMDD v3.2 and HMDD v2.0

| Method | HMDD v3.2 | | | HMDD v2.0 | | |
| --- | --- | --- | --- | --- | --- | --- |
| | AUPR | AUC | F1 | AUPR | AUC | F1 |
| NLPMMDA | 0.6564 | 0.7581 | 0.7791 | 0.6610 | 0.7635 | 0.7946 |
| CP | 0.9246 | 0.9096 | 0.8551 | 0.8419 | 0.7968 | 0.7804 |
| TFAI | 0.9248 | 0.9101 | 0.8555 | 0.8360 | 0.7912 | 0.7786 |
| **TDRC** | **0.9284** | **0.9201** | **0.8643** | **0.8663** | **0.8379** | **0.8014** |

**Table 4.** The average running time of 20 runs of different tensor decomposition methods

| Methods | Time(s) |
|---------|---------|
| **TDRC** | **13.7389** |
| CP | 31.3699 |
| TFAI | 431.6624 |

### 4.3.2 Time efficiency

We further conduct time efficiency analysis of the tensor decomposition methods. To illustrate the high-efficiency of TDRC, we use the whole ground-truth tensor in HMDD v3.2 and pre-calculated similarities as input, execute the codes of all CP-based models on a PC with 4-core i7 CPU and 24GB RAM for 20 times, and list the average running times of 20 runs in Table 4. Our proposed TDRC runs extremely faster than the other methods. CP and TFAI use the "workhorse" ALS algorithm which has to take many iterations to converge (Kolda and Bader, 2009). Moreover, at each inner alternating iteration in TFAI, it needs to solve two Sylvester equations with the size of $(m \times m)$ and $(n \times n)$ separately (recall that $m$ is the number of miRNAs and $n$ is the number of diseases), which is heavily time-consuming. Oppositely, the ADMM algorithm used in TDRC seems to be more efficient.

### 4.3 Case studies

In this part, we further evaluate the practical capability of the tensor decomposition methods for predicting unobserved miRNA-disease-type triples. We build all models by using all known four types of miRNA-disease associations in our HMDD v2.0 dataset, and then we can get predictive scores for those unknown miRNA-disease-type triples. We rank the miRNA-type pairs related with a specific disease and find evidences from HMDD v3.2 for top-ranked 20 predictions. Specifically, we here put the results for 15 selected diseases of our proposed method TDRC in Fig. 4. We can observe that the top-20 predictive precision for some diseases are no less than 50%, especially for lung neoplasms with predictive precision of 80%. The results demonstrate that tensor decomposition methods have great potential of predicting disease-associated miRNAs and the corresponding association type.

### 5 Discussion and conclusions

Predicting multiple types of miRNA-disease associations is helpful for understanding the pathogenesis of human diseases associated with the dysregulations of miRNAs. In this study, we formulate this task at the point of tensor completion and introduce a series of tensor decomposition methods for predicting unobserved miRNA-disease-type ternary associations. Further, we propose a novel tensor decomposition-based method, called TDRC, which imposes the relational constraints into the tensor decomposition model for integration of miRNA-miRNA similarity and disease-disease similarity. We provide a high-efficiency optimization algorithm for our proposed model TDRC in virtue of the ADMM framework, and resort the conjugate gradient (CG) method to avoid computing an inverse matrix in the inner iterations of ADMM for lower time complexity. Experimental results show that tensor decomposition methods overperform the baseline method. Especially, compared with other tensor-based methods, our proposed method TDRC can produce robust and satisfying performance meanwhile being efficient. Case studies also manifest the practical capability of TDRC in inferring disease-associated miRNA-type pairs for some popular diseases.

In our future work, more tensor decomposition forms will be concerned, such as Tucker decomposition (Tucker, 1966). We will also pay much attention to many similar issues that are applicable to be handled by tensor-based models, such as drug-target-disease ternary associations (Chen and Li, 2019) and multi-relational drug-drug interaction (Jin, et al., 2017).

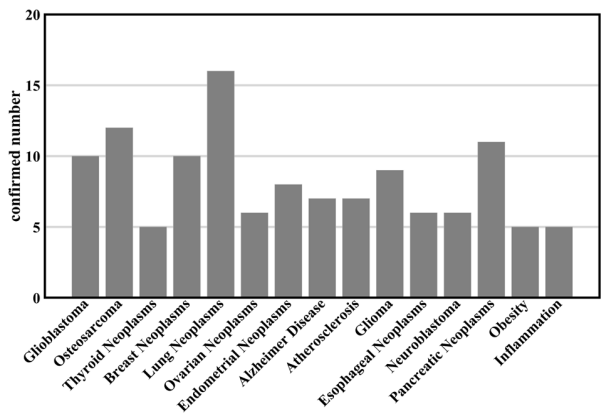

**Fig. 4. The numbers of confirmed miRNA-disease-type triplets based on HMDD v3.2 in top 20 predictions for 15 popular diseases given by TDRC based on HMDD v2.0.**

**Funding**

This work has been supported by the National Natural Science Foundation of China (61772381, 61572368), the Fundamental Research Funds for the Central Universities (2042017kf0219, 2042018kf0249), National Key Research and Development Program (2018YFC0407904), Huazhong Agricultural University Scientific & Technological Self- innovation Foundation. The funders have no role in study design, data collection, data analysis, data interpretation, or writing of the manuscript.

**References**

Bartel, D.P. (2004) MicroRNAs: genomics, biogenesis, mechanism, and function, *Cell*, **116**, 281-297.

Boyd, S., *et al.* (2011) Distributed Optimization and Statistical Learning via the Alternating Direction Method of Multipliers, *Foundations and Trends in Machine Learning*, **3**, 1-122.

Calin, G.A., *et al.* (2002) Frequent deletions and down-regulation of micro-RNA genes miR15 and miR16 at 13q14 in chronic lymphocytic leukemia, *Proceedings of the National Academy of Sciences of the United States of America*, **99**, 15524-15529.

Chen, H. and Li, J. (2019) Modeling Relational Drug-Target-Disease Interactions via Tensor Factorization with Multiple Web Sources. pp. 218-227.

Chen, X., Sun, L.-G. and Zhao, Y. (2020) NCMCMDA: miRNA-disease association prediction through neighborhood constraint matrix completion, *Briefings in bioinformatics*, 1-12.

Chen, X., *et al.* (2018) Predicting miRNA-disease association based on inductive matrix completion, *Bioinformatics (Oxford, England)*, **34**, 4256-4265.

Chen, X., *et al.* (2015) RBMMMDA: predicting multiple types of disease-microRNA associations, *Scientific reports*, **5**, 13877.

Cheng, A.M., *et al.* (2005) Antisense inhibition of human miRNAs and indications for an involvement of miRNA in cell growth and apoptosis, *Nucleic Acids Res*, **33**, 1290-1297.

Hestenes, M. and Stiefel, E.L. (1952) Methods of Conjugate Gradients for Solving Linear Systems, *Journal cf Research of the National Bureau of Standards Vol*, **49**.

Huang, Z., *et al.* (2019) Benchmark of computational methods for predicting microRNA-disease associations. *Genome Biol*. pp. 202.

Huang, Z., *et al.* (2019) HMDD v3.0: a database for experimentally supported human microRNA-disease associations, *Nucleic acids research*, **47**, D1013-D1017.

Jiang, Q., *et al.* (2009) miR2Disease: a manually curated database for microRNA deregulation in human disease, *Nucleic acids research*, **37**, D98-104.

Jin, B., *et al.* (2017) Multitask Dyadic Prediction and Its Application in Prediction of Adverse Drug-Drug Interaction. pp. 1367-1373.

Jones, K., *et al.* (2014) Plasma microRNA are disease response biomarkers in classical Hodgkin lymphoma, *Clin Cancer Res*, **20**, 253-264.

Kang, Z., *et al.* (2019) Similarity Learning via Kernel Preserving Embedding. *Proceedings of the Thirty-Third AAAI Conference on Artificial Intelligence (AAAI-19)*. pp. 4057-4064.

Karp, X. and Ambros, V. (2005) Developmental biology. Encountering microRNAs in cell fate signaling, *Science*, **310**, 1288-1289.

Kolda, T.G. and Bader, B.W. (2009) Tensor Decompositions and Applications, *SIAM Review*, **51**, 455-500.

Kossaifi, J., *et al.* (2019) TensorLy: Tensor Learning in Python, *J. Mach. Learn. Res.*, **20**, 26:21-26:26.

Li, C., *et al.* (2009) Therapeutic microRNA strategies in human cancer, *AAPS J*, **11**, 747-757.

Li, Y., *et al.* (2014) HMDD v2.0: a database for experimentally supported human microRNA and disease associations, *Nucleic Acids Res*, **42**, D1070-1074.

Liu, Y., *et al.* (2017) Inferring microRNA-disease associations by random walk on a heterogeneous network with multiple data sources, *IEEE/ACM transactions on computational biology and bioinformatics*, **14**, 905-915.

Lu, M., *et al.* (2008) An analysis of human microRNA and disease associations, *PloS one*, **3**, e3420.

Miska, E.A. (2005) How microRNAs control cell division, differentiation and death, *Curr Opin Genet Dev*, **15**, 563-568.

Narita, A., *et al.* (2012) Tensor factorization using auxiliary information, *Data Min. Knowl. Discov.*, **25**, 298-324.

Nickel, M., Tresp, V. and Kriegel, H.-P. (2011) A Three-Way Model for Collective Learning on Multi-Relational Data. pp. 809-816.

Peng, J., *et al.* (2019) A learning-based framework for miRNA-disease association identification using neural networks, *Bioinformatics (Oxford, England)*.

Rendle, S., *et al.* (2009) Learning optimal ranking with tensor factorization for tag recommendation. pp. 727-736.

Shimomura, A., *et al.* (2016) Novel combination of serum microRNA for detecting breast cancer in the early stage, *Cancer science*, **107**, 326-334.

Siebert, P.D. (1999) Quantitative rt-PCR, *Methods Mol Med*, **26**, 61-85.

Sredni, S.T., *et al.* (2011) MicroRNA expression profiling for molecular classification of pediatric brain tumors, *Pediatric blood & cancer*, **57**, 183-184.

Trouillon, T., *et al.* (2017) Knowledge Graph Completion via Complex Tensor Factorization, *J. Mach. Learn. Res.*, **18**, 130:131-130:138.

Tucker, L.R. (1966) Some mathematical notes on three-mode factor analysis, *Psychometrika*, **31**, 279-311.

Varallyay, E., Burgyan, J. and Havelda, Z. (2008) MicroRNA detection by northern blotting using locked nucleic acid probes, *Nature protocols*, **3**, 190-196.

Wang, D., *et al.* (2010) Inferring the human microRNA functional similarity and functional network based on microRNA-associated diseases, *Bioinformatics*, **26**, 1644-1650.

Xiao, Q., *et al.* (2017) A graph regularized non-negative matrix factorization method for identifying microRNA-disease associations, *Bioinformatics*.

Xu, P., Guo, M. and Hay, B.A. (2004) MicroRNAs and the regulation of cell death, *Trends in genetics : TIG*, **20**, 617-624.

Xuan, P., *et al.* (2015) Prediction of potential disease-associated microRNAs based on random walk, *Bioinformatics (Oxford, England)*, **31**, 1805-1815.

Yang, Z., *et al.* (2010) dbDEMC: a database of differentially expressed miRNAs in human cancers, *BMC Genomics*, **11 Suppl 4**, S5.

Yu, H.-F., *et al.* (2014) Large-scale multi-label learning with missing labels. *Proceedings of the 31st International Conference on International Conference on Machine Learning - Volume 32*. JMLR.org, Beijing, China, pp. I-593-I-601.

Zeng, X., *et al.* (2018) Prediction of potential disease-associated microRNAs using structural perturbation method, *Bioinformatics*, **34**, 2425-2432.

Zhang, W., *et al.* (2019) A fast linear neighborhood similarity-based network link inference method to predict microRNA-disease associations, *IEEE/ACM transactions on computational biology and bioinformatics*.

Zhang, W., *et al.* (2018) Predicting drug-disease associations by using similarity constrained matrix factorization, *BMC bioinformatics*, **19**.

Zhang, X., Yin, J. and Zhang, X. (2018) A Semi-Supervised Learning Algorithm for Predicting Four Types MiRNA-Disease Associations by Mutual Information in a Heterogeneous Network, *Genes*, **9**, 139.